\pdfoutput=1

\documentclass[11pt]{article}

\usepackage{acl}

\usepackage{times}
\usepackage{latexsym}
\usepackage{tabularx}
\usepackage{graphicx} 
\usepackage{dblfloatfix}
\usepackage[T1]{fontenc}

\usepackage[utf8]{inputenc}

\usepackage{microtype}

\usepackage{multirow}
\title{Monolingual and Cross-Lingual Acceptability Judgments \\with the Italian CoLA corpus}

\author{Daniela Trotta\textsuperscript{$\dagger$}, Raffaele Guarasci\textsuperscript{$\ddagger$}, 
\textbf{Elisa Leonardelli\textsuperscript{$\mathsection$}, Sara Tonelli\textsuperscript{$\mathsection$}} \\
  \textsuperscript{$\dagger$}University of Salerno, Italy,  \textsuperscript{$\ddagger$}ICAR-CNR, Naples, Italy \\
  \textsuperscript{$\mathsection$}Fondazione Bruno Kessler, Trento, Italy \\
  {\tt  dtrotta@unisa.it} \\
  {\tt raffaele.guarasci@icar.cnr.it} \\
    {\tt \{eleonardelli,satonelli\}@fbk.eu} \\
  }

\begin{document}
\maketitle
\begin{abstract}
The development of automated approaches to linguistic acceptability has been greatly fostered by the availability of the English CoLA corpus, which has also been included in the widely used GLUE benchmark. However, this kind of research for languages other than English, as well as  the analysis of cross-lingual approaches, has been hindered by the lack of resources with a comparable size in other languages. We have therefore developed the ItaCoLA corpus, containing almost 10,000 sentences with acceptability judgments, which has been created following the same approach and the same steps as the English one. In this paper we describe the corpus creation, we detail its content, and we present the first experiments on this new resource. We compare in-domain and out-of-domain classification,  and perform a specific evaluation of nine linguistic phenomena. We also present the first cross-lingual experiments, aimed at assessing whether multilingual transformer-based approaches can benefit from using sentences in two languages during fine-tuning.

\end{abstract}

\section{Introduction}

The ability to judge whether a sentence is perceived as natural and  well-formed by a native speaker is called acceptability judgment. Despite several open issues concerning methods for collecting and evaluating them \cite{gibson2013need,sprouse2010quantitative,Linzen2019WhatCL}, these judgments have been the most significant source of data in linguistics throughout the history of the discipline \cite{chomsky1965aspects, schutze1996empirical, naive}.

With the rise of neural language models, several works have tried to assess how much a model can encode linguistic information \cite{hewitt2019structural, manning2020emergent}, ranging from specific phenomena \cite{marvin2019targeted,goldberg2019assessing} to a general grammar knowledge \cite{Jawahar2019,mccoy2020does}. Acceptability judgments have proven to be a promising area to test the acquisition of linguistic knowledge by neural language models  \cite{gulordava2018colorless,lau-etal-2015-unsupervised}. In particular, with the creation of the Corpus of Linguistic Acceptability (CoLA) \cite{warstadt-etal-2019-neural} several approaches have been proposed that cast acceptability as a binary classification task and address it by fine-tuning transformer-based models on the corpus \cite{NEURIPS2019_dc6a7e65,DBLP:journals/corr/abs-1901-03438,JMLR:v21:20-074}. 
Unfortunately, most classification experiments on acceptability judgments have focused on English, mainly because of the lack of large  corpora in other languages.
In this work, we therefore describe the creation of a novel corpus of acceptability judgments in Italian, following the methodology used in CoLA for English. We collect 10k sentences extracted from linguistic literature and labelled by experts as acceptable or not. Furthermore, we enrich around 30\% of the sentences with additional labels describing nine linguistic phenomena. We also present a set of experiments aimed at testing the performance of a BERT-based classifier on the data and comparing it with results obtained on English. Additionally, cross-lingual experiments using XLM-RoBERTa \cite{conneau-etal-2020-unsupervised} show the potential of this approach, even if it is outperformed by monolingual models.

The main contributions of this work are therefore \textit{i)} the creation and release of the Italian Corpus of Acceptability Judgments (ItaCoLA),\footnote{Available at \url{https://github.com/dhfbk/ItaCoLA-dataset}} that to our knowledge is the largest resource of its kind in a language other than English; \textit{ii)} a set of experiments to assess the performance of BERT-based models on the whole corpus and on specific phenomena. \textit{iii)} a set of experiments using a massive multilingual language model on Italian and English, with the potential to open up novel cross-language research perspectives.

\section{Related Work}

\subsection{Acceptability corpora}

In recent years, studies on automatic assessment of acceptability have become very popular thanks to the release of the CoLa corpus \cite{warstadt-etal-2019-neural}, the first large-scale corpus of English acceptability, containing more thank 10k sentences taken from linguistic literature.

Small acceptability datasets had already been developed before, especially within the still open theoretical debate regarding the status of syntax \cite{sprouse2013empirical, lau2014measuring} and data collection methods \cite{culicover2010quantitative, gibson2013need}. These resources differ in terms of (formal or informal) data collection criteria, sources, evaluation methodology and raters used in the process.

In particular, \cite{sprouse2013assessing}, advocating the empirical status of syntax and the reliability of informal collection of acceptability judgments, test a random sample of 300 sentence extracted from the  `Linguistic Inquiry' journal. Annotators were recruited through Amazon Mechanical Turk, which had proven reliable for this type of task \cite{sprouse2011validation}. Evaluation is performed using two experimental methods: magnitude estimation and forced-choice task. These judgments are   also compared with ones collected using formal methods \cite{sprouse2013comparison} manually evaluated using a combination of AMT and naive participants without linguistic training.

 \cite{lau2014measuring} collect a dataset of 600 sentences from the BNC \cite{bnc2007british}, and then introduce infelicities using machine translation to generate sentences of varying level of grammaticality. Judgments have been collected using AMT and applying different evaluation criteria, from binary to gradient.
A recent study \cite{marvin2019targeted}, aimed at evaluating the behaviour of a neural model on specific syntactic phenomena, uses a dataset of sentence pairs automatically built using templates. 

Regarding studies on languages other than English, \cite{linzen2018reliability} evaluate informal acceptability judgments on Hebrew and Japanese collecting data from several sources ranging from peer-reviewed papers, books and dissertations. A similar study has been conducted in French \cite{feldhausen2020testing} and in Chinese \cite{chen2020assessing}. Both studies use sentences extracted from textbooks. To our knowledge, only for Swedish there is a freely-available corpus whose size is comparable to CoLA and ItaCoLA. The corpus, presented in  \cite{volodina-etal-2021-dalaj}, contains around 9,600 sentences extracted from language learner data. 

Concerning Italian, only one dataset has been released to date, in the context of Evalita 2020 evaluation campaign on complexity and acceptability (AcComplIt task) \cite{DBLP:conf/evalita/BrunatoCDMVZ20}. 
The dataset presents several differences w.r.t. ItaCoLA in terms of size, annotation approach and linguistic phenomena, which we detail in Section \ref{outofdomain}.

\subsection{Approaches to acceptability classification}
The CoLA corpus was presented together with a number of experiments aimed at assessing the performance of neural networks on a novel binary acceptability task \cite{warstadt-etal-2019-neural}. The best performance was achieved with a pooling classifier and ELMo-style embeddings, yielding 0.341 MCC on in-domain data and 0.281 on the out-of-domain test set. Matthews Correlation Coefficient (MCC) was chosen as an evaluation measure because it is more appropriate than F1 or accuracy for binary classification with unbalanced data \cite{MATTHEWS1975442}. More recently, \cite{DBLP:journals/corr/abs-1901-03438} extended the classification experiments by comparing a BiLSTM baseline with the performance achieved by transformer encoders such as GPT and BERT. The best approach is obtained by fine-tuning BERT$_{large}$ with a mean MCC of 0.582.
Other approaches, instead, focus on unsupervised learning, for example \cite{lau-etal-2015-unsupervised,lau-etal-2020-furiously} compare different types of language models to infer the probability of a sentence, which is then mapped onto acceptability.

Since CoLA has been included in the GLUE dataset \cite{wang-etal-2018-glue}, a very popular multi-task benchmark for English natural language understanding, and an acceptability challenge has been launched on Kaggle,\footnote{\url{https://www.kaggle.com/c/cola-in-domain-open-evaluation/}} the number of studies dealing with binary acceptability has remarkably increased. Unfortunately, most studies using GLUE report accuracy instead of MCC, making it difficult to identify the best approach. Nevertheless, all top-ranked systems rely on variations of transformer-based models, including ALBERT \cite{DBLP:conf/iclr/LanCGGSS20} (69.1 Accuracy) and StructBERT \cite{DBLP:conf/iclr/0225BYWXBPS20} (69.2 Acc.). More recently, also reformulating acceptability as an entailment task and using smaller language models to few-shot fine-tuning has showed a great potential \cite{wang2021entailment}, outperforming existing BERT-based approaches (86.4 Acc.).

\begin{table*}[ht]
\center
\scalebox{0.95}{
\begin{tabularx}{15cm}{|l|c|X|}
\hline
\textbf{Source}         & \textbf{Label}    & \textbf{Sentence}\\ 
\hline
\newcite{graffi94} & 0 & *Edoardo è tornato nella sua l'anno scorso città. \newline(\textit{*Edoardo returned to his last year city}) \\ 
\hline 
\newcite{graffi94} & 1 & Ho voglia di salutare Maria \newline(\textit{I want to greet Maria})\\
\hline
\newcite{graffi94} & 0 &  *Questa donna mi hanno colpito. \newline(\textit{*This woman have impressed me})  \\
\hline
\newcite{simone2013nuovi} & 1 &  Questa donna mi ha colpito. \newline (\textit{This woman has impressed me}) \\
\hline
\end{tabularx}
}
\caption{Example sentences from the ItaCoLA dataset. 1 = acceptable, 0 = not acceptable} \label{tab:sentences}
\end{table*}

Concerning acceptability on Italian, a shared task has been organised for the first time at Evalita 2020 Evaluation campaign, proposing a joint classification of complexity and acceptability \cite{DBLP:conf/evalita/BrunatoCDMVZ20}. The dataset, which we use for our out-of-domain evaluation (Section \ref{outofdomain}) was originally created merging data from different psycholinguistic studies, and includes 1,683 sentences with a manually assigned value of acceptability between 1 and 7. Two participants submitted three runs in total. In order to cope with the limited number of training instances, the best performing approach \cite{DBLP:conf/evalita/Sarti20}  implemented an ensemble of
fine-tuned models to annotate a large corpus of unlabeled text, and leveraged new annotations in a multi-task setting to obtain final
predictions over the original test set. The second system \cite{DBLP:conf/evalita/Delmonte20a} was rule-based, implementing a set of syntactic and semantic constraints to check to what an extent a sentence can be considered acceptable.

\section{ItaCoLA: Italian Corpus of Linguistic Acceptability}

In this section we introduce the Italian Corpus of Linguistic Acceptability (ItaCoLA), built with the purpose of representing a large number of linguistic phenomena while distinguishing between acceptable and not acceptable sentences. The methodology of corpus creation and its size are similar to those proposed for the English CoLA in \cite{warstadt-etal-2019-neural}, i.e. we have collected examples from different manuals covering several linguistic phenomena. This fulfills a dual purpose: the size of the corpus  allows the application of deep learning approaches to acceptability judgment, while its structure paves the way to cross-language comparative analyses. 

Concerning acceptability annotation, for the creation of ItaCoLA we have chosen to keep a  Boolean judgment in line with several previous works \cite{lawrence2000natural, wagner2009judging, linzen2016assessing}. This choice ensures robustness and simplifies classification, while allowing us to keep the original judgments as formulated by an expert (i.e. the authors of the different data sources).

\subsection{Sources}
ItaCoLA sentences come from various types of linguistic publications covering four decades. Unfortunately, the majority of linguistic textbooks or fundamental theoretical publications in Italian are not available in digital format or are not freely accessible. Therefore, the only viable way to collect data was through manual transcription.
Sources include theoretical linguistics textbooks \cite{graffi02,simone2013nuovi} and works that focus on specific phenomena such as idiomatic expressions \cite{vietri2014idiomatic}, locative constructions \cite{d1983lessico} and verb classification \cite{jezek03}.
Overall, we manually copied from a number of sources a total of 10,000 sentences, reporting also the judgment provided by the author, i.e. acceptable or not acceptable. Few examples are listed in Table \ref{tab:sentences}.

\subsection{Sentence selection}
\label{sub:selection}

Following the criteria proposed by \cite{warstadt-etal-2019-neural}, specific choices have been made to exclude some types of sentences from the corpus. This increases data consistency also for future cross-lingual experiments with CoLA. Following sentence types were not included in the dataset:

\begin{itemize}

\item Italian translations of sentences, which were originally written in other languages. The syntactic behavior of each language can cause ambiguity in judging the acceptability of translated sentences. 

\item Isolated phrases without predicative structure or full meaning expression, i.e. noun, prepositional, adjective and adverbial phrases.

\item Sentences which are difficult to evaluate without context even by a native speaker. This category includes sentences that are strictly domain-dependent, for instance statements of linguistic rules such as ``Una testa lessicale -N assegna Caso al SN che essa regge'' (En. \textit{A lexical head -N assigns Case to the SN that it holds}) or sentences extracted from novels, films or newspapers.

\item Sentences with an extremely twisted syntax and a very high number of nested subordinates. The latter are often used as borderline examples to explain phenomena such as long-distance dependencies or pro-drop, very common in Italian.

\end{itemize}

Concerning the types of sentences which have been included in the dataset, we can identify some recurring patterns. For example, there is the presence of several minimal pairs, i.e. minimally different sentences contrasting in acceptability (see the sentences in the last two rows of Table \ref{tab:sentences}). 
Other sentences are short examples created to describe or explain specific phenomena, i.e. ``Lucia lavora per studiare'' (En. \textit{Lucia works to study}) \cite{vietri2004lessico}, or elementary sentences whose syntax matches the canonical SVO (Subject-Verb-Object) order, i.e. ``Il poliziotto catturò il ladro'' (En. \textit{The policeman caught the thief}). Other common sentence types are those resulting from formal transformation tests. For instance, starting from the elementary sentence ``I bambini hanno calpestato le aiuole'' (\textit{Children stepped the flowerbeds}), other sentences can be produced by applying deletion, i.e. ``I bambini hanno calpestato'' (\textit{children stepped}), or pronominalization, i.e. ``I bambini le hanno calpestate'' (En. \textit{Children stepped on them}). These transformed sentences -- besides being in conspicuous number in the corpus -- are functional to the purpose of this work, since they are created just to verify whether native speaker intuition is validated by the data. 

\begin{table*}[ht]
\center
\scalebox{0.95}{
\begin{tabular}{|l|r|c|l|}
\hline
\textbf{Source}         & \textbf{N} & \textbf{\% acceptable} & \textbf{Topic}                   \\ \hline
\newcite{d1983lessico}        & 524        & 84.2        & locative
constructions           \\ 
\newcite{d1992analisi}        & 1,364       & 85.0        & discourse analysis               \\ 
\newcite{elia1981lessico}      & 2,167       & 84.8        & lexicon and
syntactic structures \\ 
\newcite{elia1982avverbi}               & 169        & 79.9        & locative adverbs and idioms      \\ 
\newcite{graffi02} & 157        & 84.1        & theoretical linguistics          \\ 
\newcite{graffi94}            & 604        & 79.5        & syntax                           \\ 
\newcite{graffi08}            & 122        & 82.0        & generative grammar               \\ 
\newcite{jezek03}             & 817        & 74.4        & verb
classification              \\ 
\newcite{simone2013nuovi}             & 754        & 97.7        &
theoretical linguistics          \\ 
\newcite{vietri2014idiomatic}             & 651        & 90.0        & idiomatic expressions            \\ 
\newcite{vietri2004lessico}             & 1,424       & 85.5        & lexicon-grammar approach         \\ 
\newcite{vietri2017}           & 970        & 81.4        & anticausative sentences          \\ 
\textbf{In-domain}             & \textbf{9,722}        & \textbf{84.5}        &                            \\ \hline
\end{tabular}
}
\caption{Distribution of ItaCoLA sentences by source. N is the number of sentences from each source. Topic is the main focus of the source, even if other linguistic phenomena can be present as well.}\label{tab:distribution_sources}
\end{table*}

\subsection{Data Cleaning and Refinement}
Once the sentences have been selected, some further adjustments have been made at lexical level in order to prevent possible ambiguity and make some outdated examples sound more modern. Also in this case, we follow the same principles used for CoLA. Changes have involved mainly proper nouns and verbs and have been carried out to avoid irrelevant complications due to out-of-vocabulary words:

\begin{itemize}

\item Obsolete or uncommon proper nouns and abbreviations of organisations (i.e., Ena, Isa, Lillo, Pat etc.) have been replaced when possible with more common names taken from the lists released by the Italian National Institute of Statistics.\footnote{The data consulted are updated to 2018 according to the Italian National Institute of Statistics: https://www.istat.it/it/dati-analisi-e-prodotti/contenuti-interattivi/contanomi} According to \citet{vietri2014idiomatic} mentions of rare and obsolete named entities in sentences can interfere with acceptability judgments.

\item Low-frequency terms, which in most cases pertain to the technical-specialist domain, have been manually simplified using synonyms or broader terms that made them easier to understand without affecting the semantics of the sentence. For instance the sentence ``L'artrosi ha \underline{anchilosato} le mani di Filippo.'' (En. \textit{The arthrosis has developed ankylosis Filippo's hands}) has been changed to ``L'artrosi ha \underline{paralizzato} le mani di Raffaele.'' (En. \textit{The arthrosis has paralyzed Raffaele's hands}).
\end{itemize}

In order to identify low-frequency terms, we lemmatised all sentences using the  TINT NLP suite for Italian \cite{palmero2018tint}, and then associated each lemma with the reference frequency list extracted from the Paisà corpus \cite{lyding2014paisa}. Words with a frequency < 45 were manually checked and, if possible, replaced with more frequent ones of similar meaning. In total, 130 sentences were modified in this way, while for another 17 sentences a rare word was detected but it was not possible to find a replacement without modifying the meaning of the sentence (or creating a sentence already existing in the dataset). We therefore opted for leaving these few sentences in their original form.

An additional check was performed to manually control for typos and transcription errors. We observed that some sentences were present in more than one dataset, usually in case of multiple sources by the same author.
Double sentences were thus removed (source was randomly chosen). The final dataset consists of 9,722 sentences from different sources, having each a different percentage of acceptable and not acceptable sentences, with a large prevalence of acceptable instances. An overview of the dataset is reported in  Table \ref{tab:distribution_sources}.

\begin{table*}[t]
\center
\begin{tabular}{|l|c|c|l|}
\hline
\textbf{Source}         & \textbf{N} & \textbf{\% acceptable} & \textbf{Topic}                   \\ \hline
\newcite{10.3389/fpsyg.2019.02105}            & 128        & 69.5 & object clefts          \\ 
\newcite{GRECO2020102926}             & 515        & 91.6 & copular          \\ 
\newcite{mancini18}             & 320        & 49.7 & subject-verb agreement          \\ 
\newcite{villata2015intervention}            & 48        & 66.7 & wh-violations          \\ 
\newcite{chowdhury-zamparelli-2018-rnn}          & 672        & 53.6 & various from templates          \\ 
\textbf{Out-of-domain}             & 1,683       & 66.0      &                            \\ \hline

\end{tabular}
\caption{The content of AcComplIt dataset \cite{DBLP:conf/evalita/BrunatoCDMVZ20} used for out-of-domain experiments}\label{tab:evalita}
\end{table*}

\section{Monolingual Experiments}
The monolingual experiments are aimed at presenting the first classification results on ItaCoLA and at defining standard training, validation and test split, to be used also in future experiments with the corpus.  
We compare two classifiers: one using LSTM and FastText embeddings, which we consider our baseline, and the other using an Italian version of BERT \cite{devlin-etal-2019-bert},  which we fine-tune using ItaCoLA training dataset.
The two classifiers are evaluated in an  in-domain and an out-of-domain setting, similar to the evaluation performed on English CoLA.  \cite{warstadt-etal-2019-neural}. 

For the \textbf{in-domain evaluation}, we divide the ItaCoLA corpus into a training, a validation and a test split, including respectively 7,801,  946 and 975 examples. We create the splits so that each source is equally represented in each  split and the acceptability/not acceptability ratio is preserved.
For the \textbf{out-of-domain setting}, training is performed on the same  split used for the in-domain experiments. Validation and test, instead, are carried out using the AcComplIt dataset \cite{DBLP:conf/evalita/BrunatoCDMVZ20}. In particular, for validation we use the training set released for the Evalita shared task and for testing we use the official AcComplIt test set. We consider this dataset out-of-domain not only because it comes from different sources compared to ItaCoLA, but also because it was created using crowd-sourcing, i.e. following a completely different approach than ours, which relies on linguistic literature.

\par \textbf{Baseline LSTM}: 
As baseline classifier, we implement a bidirectional LSTM with two layers (64 and 32 neurons) and a dropout of 0.3. Each sentence is represented as a sequence of word embeddings, obtained with the Italian model of FastText \cite{grave2018learning} trained on Common Crawl and Wikipedia with size 300.\footnote{\url{https://github.com/facebookresearch/fastText/blob/master/docs/crawl-vectors.md}} The network is implemented with Keras \cite{franoischollet2017learning} (Adam optimizer, learning rate 0.01, loss function: binary crossentropy, 15 epochs). We perform 10 restarts.  Reported results represent the mean performance obtained over the restarts.

\par \textbf{BERT}: Among the Italian BERT-like versions available, we select \emph{Bert-base-italian-xxl-cased}, available on Huggingface.\footnote{\url{https://huggingface.co/dbmdz/bert-base-italian-xxl-cased}} It is a model pre-trained on a total general-purpose corpus of 81GB. 
After randomizing the order of instances in our training set, we fine-tune the model using PyTorch,\footnote{\url{ https://pytorch.org/}} with a maximum sequence length of 64, a batch size of 32 for 12 epochs. We perform 10 restarts. Also in this case, reported results are the mean across the repeated classifications.

\begin{table*}[ht]
\center
\scalebox{0.95}{
\begin{tabular}{|l|c|c|c|c|}
\hline
\multirow{2}{*}{\textbf{Model}} & \multicolumn{2}{c}{\textbf{In-domain}} & \multicolumn{2}{|c|}{\textbf{Out-of-domain}} \\ \cline{2-5}
        &  {Acc.} & {MCC} & {Acc.} & {MCC} \\ \hline

LSTM &0.794 &0.278
\small{$\pm$ 0.029} \normalsize{(best: 0.334)} &0.605 & 0.147 
\small{$\pm$ 0.066} \normalsize{(best: 0.213)}  \\ \hline
ITA-BERT & 0.904 &0.603 \small{$\pm$ 0.022} \normalsize{(best: 0.627)}
 & 0.683 & 0.198 \small{$\pm$ 0.036} \normalsize{(best: 0.255)}  \\ \hline
\end{tabular}
}
\caption{Classification results on the ItaCoLA test set and the out-of-domain AcComplIt test set. Results are the mean of 10 runs $\pm$ StdDev. Best result between parenthesis.
}\label{tab:results}
\end{table*}

\subsection{In-domain results}
Results on the in-domain test set are displayed in Table \ref{tab:results}. We report both Matthews Correlation Coefficient (MCC) \cite{peters2018deep}, which is the score originally proposed by CoLA authors, and Accuracy. MCC  is a measure of correlation for Boolean variables and it is particularly suited when evaluating unbalanced binary classifiers. We report Accuracy as well, which is instead generally used to evaluate acceptability on the GLUE benchmark.
Classification performance is in line with the results obtained for English, since \citet{DBLP:journals/corr/abs-1901-03438} report MCC = 0.582 (mean of 20 restarts) using BERT$_{large}$ and MCC = 0.320 with the LSTM baseline on in-domain data. 
In general, these results suggest that neural approaches applied to Italian can work with a performance similar to English, provided that the same amount of training data is available.

\subsection{Out-of-domain results}
\label{outofdomain}
Since acceptability in the AcComplIt dataset used for out-of-domain evaluation is labelled for perceived acceptability on a 7-point Likert scale, we first map these labels to two classes (i.e. acceptable or not) if the average score is $\geq$ 3.5 or below, respectively. We report statistics related to the composition of the dataset and the distribution of acceptable sentences in Table \ref{tab:evalita}

Also in this case classification results are reported in Table \ref{tab:results}. Similar to the in-domain data, the BERT-based classifier outperforms the LSTM baseline. However, results are much lower than those reported in the same setting for English, where the best result obtained with a pooling classifier and ELMo-style embeddings is MCC = 0.281. This difference is probably due to a number of factors, including the different approach followed to create the out-of-domain dataset, the fact that we mapped the Likert scale into two classes, and the presence of different linguistic phenomena. Another difference is the average sentence length: while it is 6 tokens in ItaCoLA, sentences in AcComplIt contain on average 10 tokens. Furthermore, in AcComplIt the percentage of not acceptable sentences is higher than in ItaCoLA, i.e. 24\% vs. 16\% respectively.

\section{Analysis of Specific Linguistic Phenomena}
Acceptability judgments involve a number of different linguistic phenomena, which we tried to cover as much as possible by selecting different sources for the creation of the dataset. However, in order to fully understand how well classifiers can judge acceptability in the presence of these phenomena, we perform also a fine-grained evaluation focused on specific linguistic constructions. 

\subsection{Data Annotation}
\label{sub:finegrained}
We annotate a subset of the corpus with nine linguistic phenomena. The sentences to be annotated have been selected by manually going through the dataset and extracting examples showing at least one of the phenomena of interest, until around 20\% of the overall dataset was annotated. In total 2,088 sentences were annotated, with 2,729 phenomena (1.3 average phenomenon per sentence). 

The annotated phenomena can be divided in two macro-groups. The first one contains roughly the same classes proposed for  the AcComplit dataset \cite{DBLP:conf/evalita/BrunatoCDMVZ20}, which we use for our out-of-domain evaluation. These classes are reported below as items 1 -- 4. The second set of phenomena (items 5 -- 9) includes some of the traits annotated in  \citet{DBLP:journals/corr/abs-1901-03438} for the English language, although it is not always possible to guarantee perfect equivalence between the syntax of the two languages. 
We detail them as follows:

\begin{itemize}

\item[1)] \textbf{Cleft constructions} 
(136 sentences): Sentences where a constituent has been moved to put it in focus, e.g. ``È il toro che Aurora ha preso per le corna e non il bufalo'' (En. \textit{It is the bull that Aurora has taken by the horns and not the buffalo}.) 

\item[2)] \textbf{Copular constructions} 
(855 sentences): Sentences with a copulative verb that joins the subject of the sentence to a noun or an adjective, e.g. ``Francesco è un grande oratore'' (En. \textit{Francesco is a great speaker}.) 

\item[3)] \textbf{Subject-verb agreement} 
(406 sentences): Sentences characterized by the presence or lack of subject and verb agreement in gender or number, e.g. ``Lorenzo ha detto che Andrea ha parlato con Riccardo'' (En. \textit{Lorenzo said that Andrea talked to Riccardo}.)

\item[4)] \textbf{Wh-islands violations} 
(53 sentences): Sentences introduced by a Wh- clause presenting correct or wrong syntactic constructions, e.g. ``Che libro dice che il professore ha raccomandato di leggere?'' (En. \textit{What book does the professor say he recommended you to read?})

\item[5)] \textbf{Simple} 
(365 sentences): Sentences in which only one verb and the mandatory arguments  are present, e.g. ``Tommaso legge il giornale''  (En. \textit{Tommaso reads the newspaper.})

\item[6)] \textbf{Question} 
(177 sentences): Interrogative sentences,  
e.g. ``Chi mi ha colpito?''  (En. \textit{Who hit me?})

\item[7)] \textbf{Auxiliary} 
(398 sentences): Sentences containing one  of the two auxiliary verbs in Italian, i.e. ``essere'' (\textit{to be}) or ``avere'' (\textit{to have}), e.g. ``Sono arrivati molti ragazzi'' (En. \textit{A lot of guys came in.})

\item[8)] \textbf{Bind} 
(27 sentences): Sentences that contain free pronouns, generally used in Italian to create contrast or focus when used together with the intensifier ``stesso'' (\textit{itself}), e.g. ``Lorenzo allieta se stesso'' (En. \textit{Lorenzo cheers himself.})

\item[9)] \textbf{Indefinite pronouns} 
(312 sentences): Sentences containing pronouns that indicate someone or something in a generic and indefinite way, e.g. ``Cerco qualcuno con cui parlare'' (En. \textit{I'm looking for someone to talk to.})

\end{itemize}

\subsection{Evaluation}
To obtain a better insight into classifier performance on different linguistic phenomena, we evaluate the Italian BERT model also in this setting. To this purpose, we modify the train/test/validation splits: all 2,088 sentences annotated with fine-grained phenomena are used as test set, while the remaining part of the dataset (7,632 sentences) is used for training (6,833 sentences) and validation (800 sentences).
We fine-tune \emph {Bert-base-italian-xxl-cased} with the same parameters reported for the previous experiments. Also in this case we perform 10 restarts.  

\begin{figure*}
    \centering
    \includegraphics[scale=0.26]{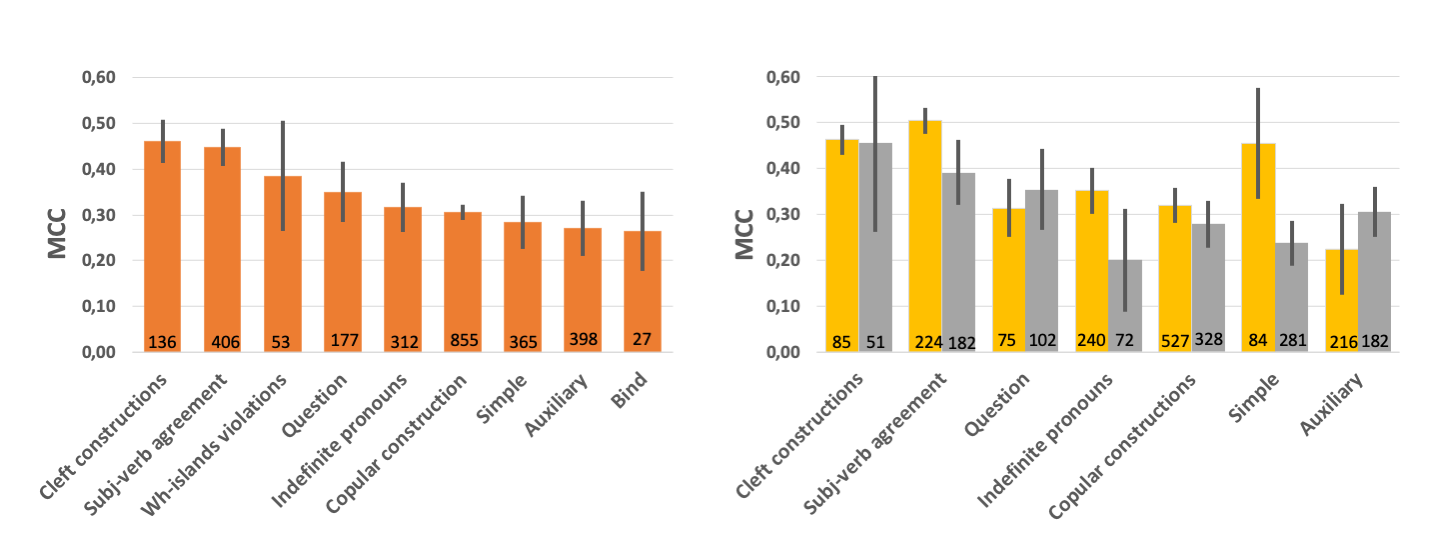}
    \caption{Classification results on a subset of ItaCoLA sentences annotated with different linguistic phenomena. Overall performance (left) and classifier performance distinguishing between sentences showing only one phenomenon (yellow) and multiple ones (grey). The number at the bottom of the bar corresponds to the number of test sentences for each phenomenon. The Bind class has been removed from the right chart because it includes only 27 sentences.}
    \label{fig:phenomenon}
\end{figure*}

Results are reported in Fig. \ref{fig:phenomenon} (left).
Overall, we observe a high variability across different phenomena. Some constructions seem to be easier to handle such as Clefts and Subject-Verb Agreement. Surprisingly, Simple sentences do not achieve the highest results despite their linear syntax, which reflects the dominant SVO word order in Italian \cite{liu2010dependency}. On English, instead,  \newcite{DBLP:journals/corr/abs-1901-03438} report for this category the best classification results in CoLA.  Another evident difference between the two languages is that Copula constructions and Wh-violations are classified poorly in Italian, while \citet{DBLP:journals/corr/abs-1901-03438} report for both MCC > 0.50.

Results on Italian are probably influenced by the presence of multiple phenomena in the same sentence. Indeed, 29\% of the sentences bears multiple annotations. As regards Simple sentences, we hypothesize that they tend to be wrongly classified because of the presence of other linguistic phenomena among the ones considered: only 23\% of the Simple sentences in our sample have not been annotated with another label. 

By re-running classification only on this subset, we observe indeed that performance increases up to 0.455 MCC. The fact that classification of sentences containing only one phenomenon yields better results holds for all categories except for Questions and Auxiliary. We report in Figure \ref{fig:phenomenon} (right) a detailed analysis of classification performance distinguishing between sentences with only one label (yellow bars) and with multiple annotated phenomena (grey  bars). Interestingly, Wh-islands violation does not appear in the chart on the right because this phenomenon is always accompanied with at least another annotation. MCC on sentences with single labels is on average 0.363 $\pm$ 0.021, while it drops to 0.308 $\pm$ 0.041 for sentences with multiple annotated phenomena.

\section{Cross-lingual Experiments}

Given that ItaCoLA has been created following the same principles of English CoLA and that monolingual results on Italian are in line with results obtained on the English dataset using a similar BERT-based approach, we perform a first set of cross-lingual classification experiments, to serve as baseline results for future improvements. We rely on XLM-RoBERTa-base \cite{conneau-etal-2020-unsupervised}, a large multi-lingual language model, trained on 2.5TB of filtered CommonCrawl data. We experiment with  different classification settings, which are all evaluated both on ItaCoLA and on CoLA in-domain test sets. This means that, starting from the same multilingual model, we classify English and Italian sentences. We implement the model in Pytorch, using a batch size of 32 and a max sequence length of 64. The learning rate is set to 2e-5, and training goes for 12 epochs. Three restarts are performed for each experiment. The number of restarts was constrained by the fact that evaluation of the English test set was possible only through Kaggle, which limits the number of runs that can be submitted for evaluation.
Results are reported in Table \ref{tab:multilingual}. We compare three models: one obtained by fine-tuning XLM-RoBERTa with English and Italian training set together, one using only the English training, and one using only Italian sentences. Each model is tested on both languages separately. 
Results show that in this setting cross-lingual zero-shot learning still performs poorly  (MCC = 0.114 both for English and Italian). When training using both languages, results outperform training and testing on the same language, showing the potential of this approach.  
However, results obtained using XLM-RoBERTa are largely outperformed by the monolingual BERT model (Table \ref{tab:results}), confirming the findings already reported in studies on other NLP tasks \cite{DBLP:journals/corr/abs-2003-02912}. 

\begin{table*}[ht]
\center
\scalebox{0.95}{
\begin{tabular}{|c|c|c|c|c|}
\hline
 \multirow{2}{*}{\textbf{Training and validation}}  &  \multicolumn{2}{c}{\textbf{Test: ItaCoLA}} & \multicolumn{2}{|c|}{\textbf{Test: CoLA}} \\ \cline{2-5}
        &    { Acc. }& {MCC} & { Acc.* }& {MCC} \\ \hline

 ItaCoLA and CoLA &0.88 &0.517 \small{$\pm$ 0.044} \normalsize{(best: 0.553)} &  0.82 & 0.508 \small{$\pm$ 0.029} \normalsize{(best:0.535)}  \\ \hline
only CoLA &0.82 &0.114 \small{$\pm$ 0.027} \normalsize{(best:0.142)} & 0.81 &
0.453 \small{$\pm$ 0.04} \normalsize{(best:0.494)}

\\ \hline
only ItaCoLA & 0.86 & 0.440 \small{$\pm$ 0.054} \normalsize{(best: 0.497)} & 0.76 & 0.114 \small{$\pm$ 0.136} \normalsize{(best:0.211)} \\ \hline
\end{tabular}
}

\caption{Monlingual and cross-lingual classification results using XLM-RoBERTa. MCC is the average of three restarts $\pm$ StdDev.
*For CoLA accuracy is calculated on development set, while MCC on test set via Kaggle because the test set is not available.}\label{tab:multilingual}
\end{table*}

\section{Conclusions}
In this paper we present the Italian Corpus of Linguistic Acceptability, a novel dataset including almost 10k sentences taken from different linguistic resources with a binary annotation of acceptability. The corpus is released in three splits (training, development and test set) so to make replicability and further experiments easier. Part of the dataset has also been manually annotated with 9 linguistic phenomena, enabling a fine-grained evaluation of the classifier performance on specific dimensions. The process to create the corpus has followed as much as possible the one adopted to collect the English CoLA, which has become the \textit{de facto} standard dataset for linguistic acceptability and has greatly fostered the development of automated systems for acceptability judgments. ItaCoLA can represent a first step towards the creation of multilingual benchmarks for acceptability, in line with recent efforts to create massive multilingual resources covering different tasks \cite{hu2020xtreme}. 

In the future, we plan to further explore the differences between ItaCoLA and AcComplit \cite{DBLP:conf/evalita/BrunatoCDMVZ20}, the other existing dataset for acceptability in Italian. We will also experiment with the Swedish corpus for acceptability studies presented in  \newcite{volodina-etal-2021-dalaj}, to check whether the findings in our work, in particular the cross-lingual experiments, hold also for Swedish when paired with English and Italian. Furthermore, we plan to explore classification approaches that yield state-of-the-art results on CoLA. While some of them are not applicable to the new corpus, because of the lack of many massive LMs for Italian, recent studies showed that with smaller language models it should be possible to achieve better results after reformulating NLP tasks as entailment \cite{wang2021entailment}. We will explore whether this research direction is promising also for acceptability studies for languages with limited resources.

\bibliography{acl_latex}
\bibliographystyle{acl_natbib}

\end{document}